\documentclass{ecai}
\usepackage{times}
\usepackage{graphicx}
\usepackage{latexsym}

\begin{document}
  \title{
    Uncovering the Data-Related Limits of Human Reasoning Research:\\
    An Analysis based on Recommender Systems
  }

  \author{
    Nicolas Riesterer \and Daniel Brand \and Marco Ragni\institute{Cognitive Computation Lab, University of Freiburg, email: riestern@cs.uni-freiburg.de}
  }

  \maketitle
  \bibliographystyle{ecai}

  \begin{abstract}
    Understanding the fundamentals of human reasoning is central to the development of any system built to closely interact with humans. Cognitive science pursues the goal of modeling human-like intelligence from a theory-driven perspective with a strong focus on explainability. Syllogistic reasoning as one of the core domains of human reasoning research has seen a surge of computational models being developed over the last years. However, recent analyses of models' predictive performances revealed a stagnation in improvement. We believe that most of the problems encountered in cognitive science are not due to the specific models that have been developed but can be traced back to the peculiarities of behavioral data instead.

    Therefore, we investigate potential data-related reasons for the problems in human reasoning research by comparing model performances on human and artificially generated datasets. In particular, we apply collaborative filtering recommenders to investigate the adversarial effects of inconsistencies and noise in data and illustrate the potential for data-driven methods in a field of research predominantly concerned with gaining high-level theoretical insight into a domain.

    Our work (i) provides insight into the levels of noise to be expected from human responses in reasoning data, (ii) uncovers evidence for an upper-bound of performance that is close to being reached urging for an extension of the modeling task, and (iii) introduces the tools and presents initial results to pioneer a new paradigm for investigating and modeling reasoning focusing on predicting responses for individual human reasoners.
  \end{abstract}

  \section{Introduction}
      The goal of human-level AI is currently approached from two directions: solving tasks with a performance similar to or even exceeding the one of humans \cite{He2015}, and understanding human cognition to a level that allows for an application to real-world problems \cite{Timm2018}. While the first direction has seen major progress mainly fueled by the development of high-performant data-driven methods over the course of the last years, the second lags behind.

Gaining insight into the processes underlying human cognition is the core focus of cognitive science, the inter-disciplinary research area at the junction of artificial intelligence, cognitive psychology, and neuroscience. Currently, this field of research is focused mainly on the psychological questions related to cognition. As a consequence, there is a distinct lack of readily available computational models developed for use in real-world applications such as human-like assistant systems. In this article we propose the use of methods from information retrieval to perform data analyses for investigating the remaining potential in modeling human cognition. In particular, we apply models from the family of collaborative filtering recommendation systems (for introduction see \cite{Sarwar2001,Ricci2010}) to re-evaluate the theory-focused state of the art and illustrate the potential of a more data-driven approach to modeling human reasoning in one of its core domains: syllogistic reasoning.

Syllogisms are one of the core domains of human reasoning research. They are concerned with categorical assertions of the form ``All A are B; All B are C'' consisting of two premises featuring a quantifier out of ``All'', ``Some'', ``Some not'', and ``No'', and three terms, A, B, and C, two of which are uniquely tied to their respective premise. Depending on the arrangement of terms, the syllogism is said to be in one of four figures (\emph{A-B;B-C}, \emph{B-A;C-B}, \emph{B-A;B-C}, \emph{A-B;C-B}).

When presented with syllogistic problems, the goal is to determine the logically valid conclusion out of the nine possibilities constructed by relating the two end terms of the premises via one of the four quantifiers (eight options), or to respond with ``No Valid Conclusion'' (NVC) if nothing else can be concluded. Featuring 64 distinct problems with nine conclusion options, the domain is well-defined, small enough to gain interpretable insight, but more detailed than most of its alternatives such as conditional reasoning (``If it rains, then the street is wet; It rains'').

The domain of human syllogistic reasoning has seen an increase of interest in modeling over the last years. A meta-analysis compiled a list consisting of twelve accounts trying to provide explanations for the behavior of humans which differs drastically from formal logics \cite{Khemlani2012}. However, since cognitive science follows a strongly theory-driven perspective on modeling, the focus of interest often rests on analyzing and comparing specific properties of models instead of their general predictive performance. Recent work identified a lack of predictive accuracy of cognitive models which raises concerns about their general expressiveness \cite{Riesterer2018b}.

In this article, we briefly analyze the predictive accuracy of the state of the art in modeling human syllogistic reasoning and compare the results with data-driven models. In particular, we apply collaborative filtering-based recommender systems which exhibit properties making them promising tools for cognitive research. We leverage these properties to test structural assumptions about the syllogistic domain to analyze the data's information content and the impact of noise on model performance. Finally, the implications for modeling reasoning and working with human data in general are discussed and ideas for improving the cognitive modeling problem are proposed.




  \section{Related Work}
      Computational modeling has become one of the prime choices for formalizing knowledge and understanding about a domain of interest. By implementing intuition and assumptions into computationally tractable models, competing theories can be evaluated, progress in the understanding of a domain can be monitored, and finally, real-world applications can be solved \cite{Newell1973}.

The field of syllogistic reasoning has seen a rise of computational models. From initially only verbally described abstract theories \cite{Sun2009}, a recent meta-analysis compiled a list of twelve theoretical accounts for syllogistic reasoning, seven of which could be specified via tables relating syllogistic problems with sets of possible conclusions \cite{Khemlani2012}. While these prediction tables still are far off fully specified implementations of the theoretical foundations, they can serve as a starting point for conducting model evaluation and comparison. The authors of the meta-analysis used the prediction data in order to determine strengths and weaknesses of the competing approaches when compared to dichotomized human response data via classification metrics (hits, misses, and false alarms). They found that while the approaches all exhibit distinct properties with respect to predictive precision, no single model could be determined as an overall winner.

A recent analysis focusing on combining individual models' strengths while avoiding their weaknesses took the evaluation of models one step further by avoiding the data aggregation step and focusing on the performance obtained from querying models for individual response predictions instead \cite{Riesterer2018b}. Their work revealed substantial lack of predictive performance of state-of-the-art models for syllogistic reasoning. Simultaneously, the authors demonstrated that data-driven modeling in form of a predictor portfolio, could be applied successfully to increase the predictive accuracy on the task.

Information systems and machine learning as the fields concerned with data-driven model construction and optimization have seen an astonishing increase in popularity over the last years. Parts of this success have been due to an integration of features related to personalities of individuals \cite{Pennock2000}. Still, even though they share methods such as clustering, principle component analyses or mixed models, they have yet to enter the domain of cognitive research.
Collaborative filtering as one of the default methods in the field of recommender systems has been successfully applied to model human reasoning before \cite{Kola2018}. What makes this kind of memory-based collaborative filtering approaches promising for cognitive research in general is their high predictive capabilities paired with the similarity to the core assumption of cognitive science, that groups of people share similar reasoning patterns. Since recommendations are extracted from similarities between different features of the data or the users themselves, they allow both for an analysis of the data underlying the recommendation process, and an analysis of high-level theoretical assumptions which can be integrated directly into the model's algorithmic structure (e.g., the integration of user personality \cite{Pennock2000,Nalmpantis2017,Hu2011}).

The following sections contrast the models from cognitive science with collaborative filtering-based approaches in a general benchmarking setting for syllogistic reasoning based on predictive accuracy.

  \section{Benchmarking Syllogistic Models}

\begin{figure}
    \centering
    \includegraphics[width=0.65\columnwidth]{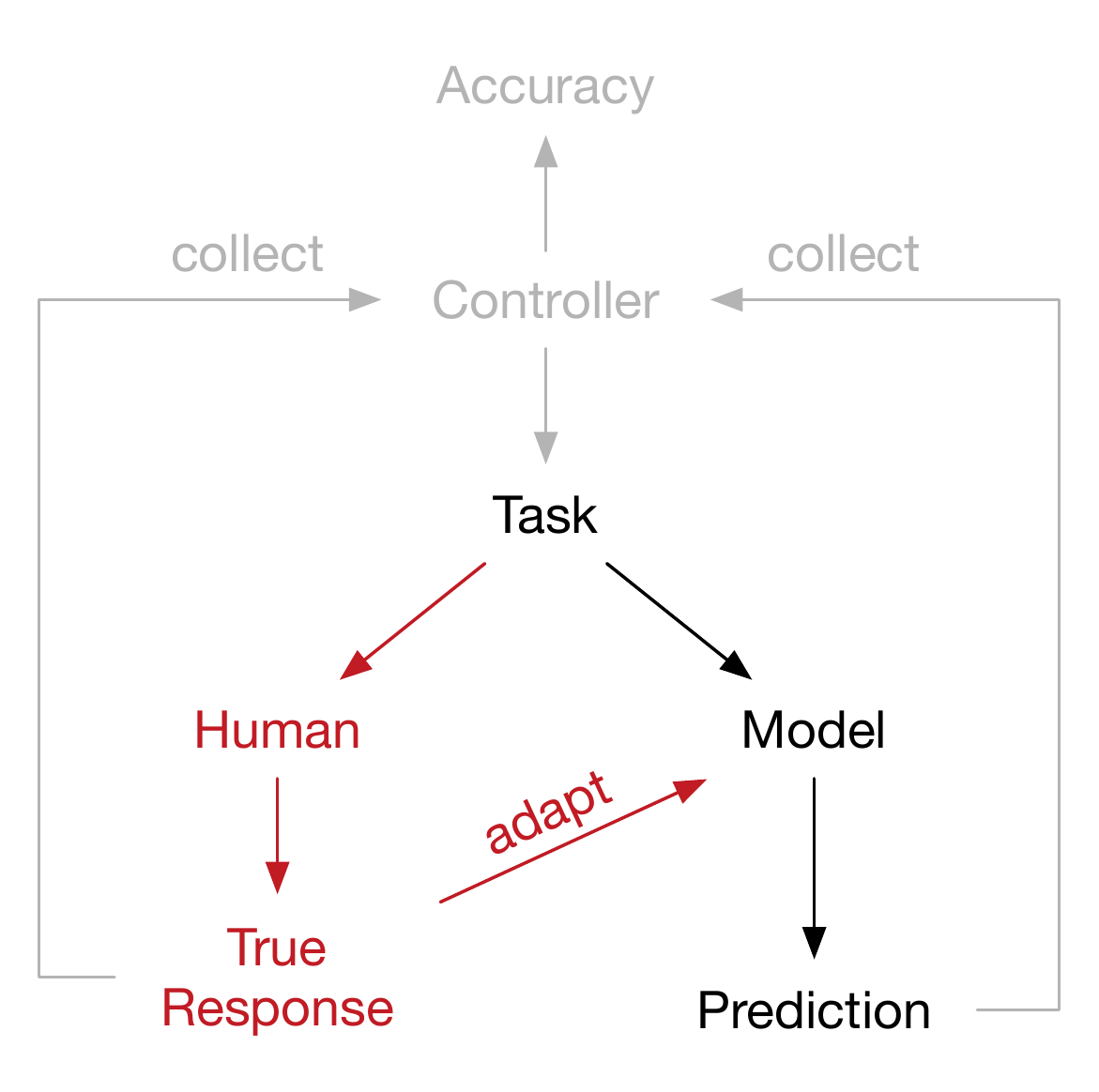}
    \caption{Overview over the model evaluation procedure. The benchmark selects a task which is fed to the model in order to obtain a prediction (black arrows). Simultaneously, by being based on experimental data, it simulates querying a human for a response (red arrows). After obtaining the model prediction, the true response is revealed to the model in an adaption step. The true (human) and model conclusions are collected and ultimately evaluated in terms of predictive accuracy.}
    \label{img:ccobra-flow}
\end{figure}


    To gain an overview over the state-of-the-art's performance in the prediction task, we performed a benchmark analysis using data obtained from an online experiment conducted on Amazon Mechanical Turk consisting of $139$ reasoners which responded to all $64$ syllogisms. Evaluations were computed relying on leave-one-out crossvalidation, i.e., by testing one reasoner and supplying the remaining $138$ as training data.

    The model evaluation procedure is inspired by a live prediction scenario where model predictions are retrieved simultaneously to the human reasoner selecting a conclusion. This is illustrated by Figure~\ref{img:ccobra-flow}. In particular, our benchmark simulates this experiment by passing the tasks to a model generating predictions (black arrows). After a prediction is obtained, the model is supplied with the true response obtained from the human reasoner (red arrows). This allows models to perform an adaptation to an individual's reasoning processes. Predictions and true responses are collected and finally compared to compute the predictive accuracy as the average number of hits.

    We included the cognitive models (matching, atmosphere, probability heuristics model, PHM; mental models theory, MMT; PSYCOP, conversion, verbal models) supplied with the meta-analysis on syllogistic reasoning by extracting the prediction tables \cite{Khemlani2012}. Additionally, we included two baseline models, \emph{Random} and \emph{MFA}. Random represents a lower bound of predictive performance defined by the strategy that always picks a random response out of the nine options. MFA denotes the most-frequent answer strategy which generates predictions by responding with the conclusion most frequently occurring in the training data. Finally, we included two variants of memory-based collaborative filtering. The user-based variant (UBCF) generates its prediction based on the responses of other users weighted by the similarity computed as the number of matching responses. The item-based variant (IBCF) compiles an item x item matrix $\mathbf{M}$ of corresponding responses (i.e., who responded with $x$ to syllogism $A$ also responded with $y$ to $B$) and a user vector $\mathbf{u}$ consisting of the user's previous responses. The prediction is generated by selecting the highest-rated response for a syllogism from the result of the matrix-vector multiplication $\mathbf{M} \times \mathbf{u}$.

    \begin{figure*}
        \centering
        \includegraphics[width=0.8\textwidth]{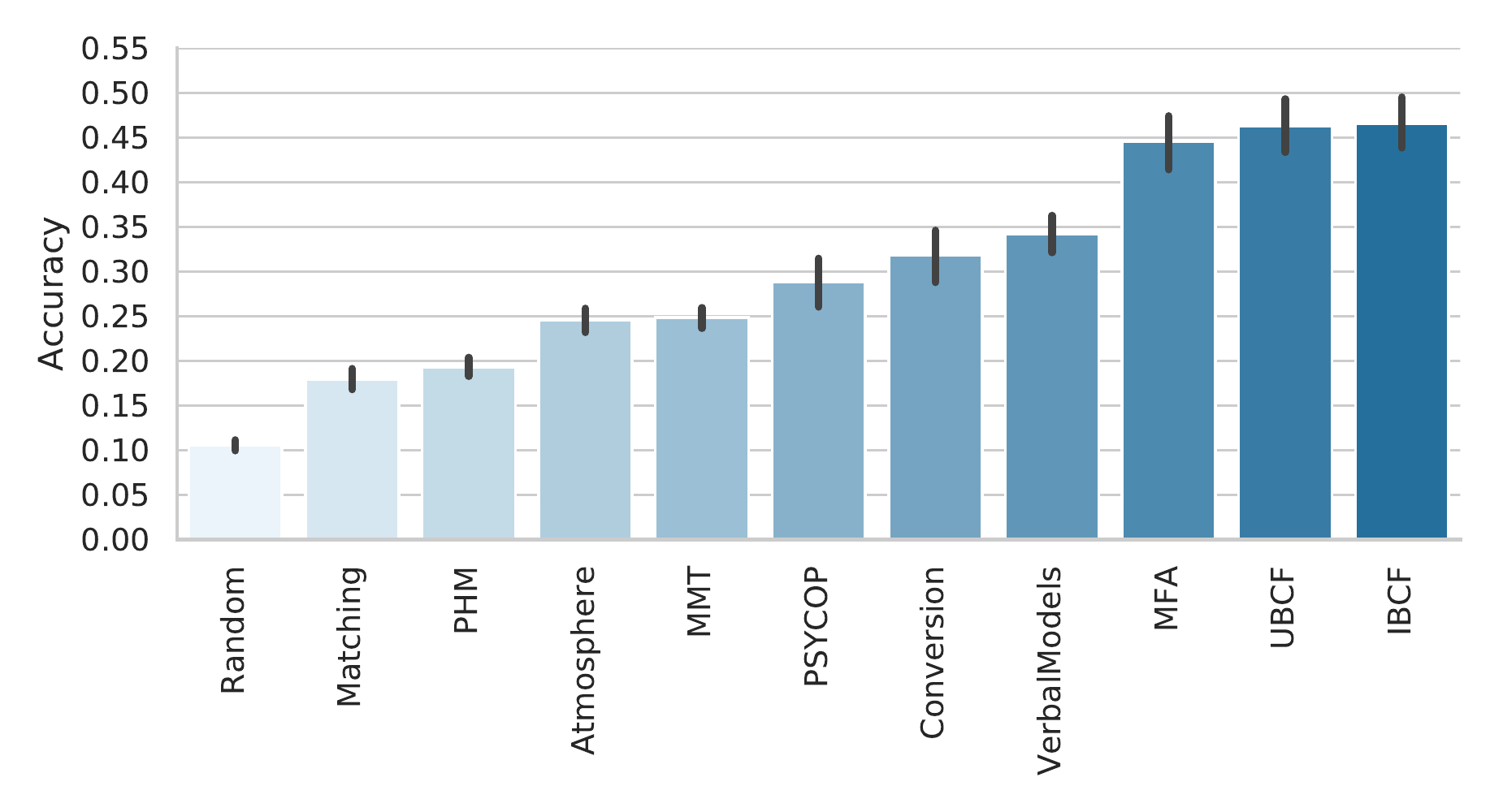}
        \caption{Accuracies of models for human syllogistic reasoning. The plot includes cognitive models based on prediction tables reported by a recent meta-analysis by Khemlani \& Johnson-Laird (2012; Probability Heuristics Model, PHM; Mental Models Theory, MMT; Matching, Atmosphere, PSYCOP, Conversion, VerbalModels), baseline models (Most Frequent Answer, MFA; Random), as well as user-based collaborative filtering (UBCF) and item-based collaborative filtering (IBCF).}
        \label{img:ccobra-benchmark}
    \end{figure*}

    Figure~\ref{img:ccobra-benchmark} depicts the result of the benchmark analysis. The image highlights the difference between cognitive models and the recommenders. This is not too surprising since most cognitive models were not introduced with predictive performance in mind. They were originally based on some statistical effect (e.g., illicit conversion, a bias towards misinterpreting the direction of the input premises \cite{Chapman1959}) or a high-level cognitive theory (e.g., PSYCOP which assumes that reasoning is the result of interactions between different mental rules \cite{Rips1994}) and are analyzed with respect to their qualities in reproducing aggregate effects of data. Still, the gap between cognitive models and data-driven approaches calls for a re-thinking of the goals of cognitive science. If the high-level insight cannot be integrated into successful models, their analysis is of limited use for advancing the understanding of human cognition.

    When observing the plot, special emphasis should be placed on MFA, the baseline model responding with the most-frequent answer of the training dataset. In terms of data-driven approaches, the MFA represents an upper bound of performance for models which do not take inter-individual differences into consideration. Since the cognitive models we considered for our analysis lack computational mechansisms for handling differences between reasoners, they are not expected to score higher than the $45\%$ achieved by MFA. In general, models can only hope to score higher if they rely on an active adaption to information about an individual's reasoning processes such as previous responses or other personality traits known to influence cognition such as working memory capacity \cite{Sues2002}.

    Being defined on an explicit database of information, collaborative filtering is an ideal tool for data analysis and modeling. They allow researchers to directly incorporate knowledge about the domain into the recommendation process and thereby to directly evaluate the value of findings in rigorous modeling scenarios. However, since this transformation of abstract findings is out of scope for this article and remains a challenge for future research, we do not focus on proposing an optimal recommender. We rather intend to highlight the method's potential for future research in the domain by illustrating the levels of performance standard domain-agnostic implementations can achieve.

    Our benchmark shows that even domain-agnostic recommenders outperform cognitive models. Still, they do not manage to significantly surpass MFA. This could mean (i) that these models fail to recognize the reasoning strategies underlying the data, or (ii) that human reasoning is too irregular, i.e., too prone to uncontrollable noise for the approaches to succeed. In the following section we analyze artificially generated data in order to gain further information about the reasons behind the limited predictive performance of syllogistic models.

  \section{Simulation Analysis}
      
    A core assumption of cognitive science is that reasoning is the result of different processes \cite{Evans2011}. Depending on the individual state of the reasoner (e.g., previous experience or concentration), thorough inferences based on the rules underlying formal logics can be conducted or simple heuristic rules can be applied to reach a conclusion. Consequently, when assessing reasoning data, it is usually assumed that the data at hand is the result of multiple interleaved strategies which need to be disentangled in order to allow for an interpretable analysis.

\subsection{Entropy Analysis}

    High information content in data is essential for the success of data-driven methods. If the data consists mostly of random effects with little structure, patterns cannot be recognized to base future predictions on.

    A common measure of information is the Shannon entropy $S$:

    \[S = -\sum_i p_i \log_2 p_i\]

    Entropy can be understood as a measure of unpredictability of a state defined via the probabilities $p_i$. In the case of syllogisms, entropy has previously been applied to quantify the difficulty of the 64 problems \cite{Khemlani2012}. Higher entropy results from a more uniform spread of probability mass over the nine conclusion options and thus serves as an indicant for a more difficult task.

    \begin{figure}
        \includegraphics[width=\columnwidth]{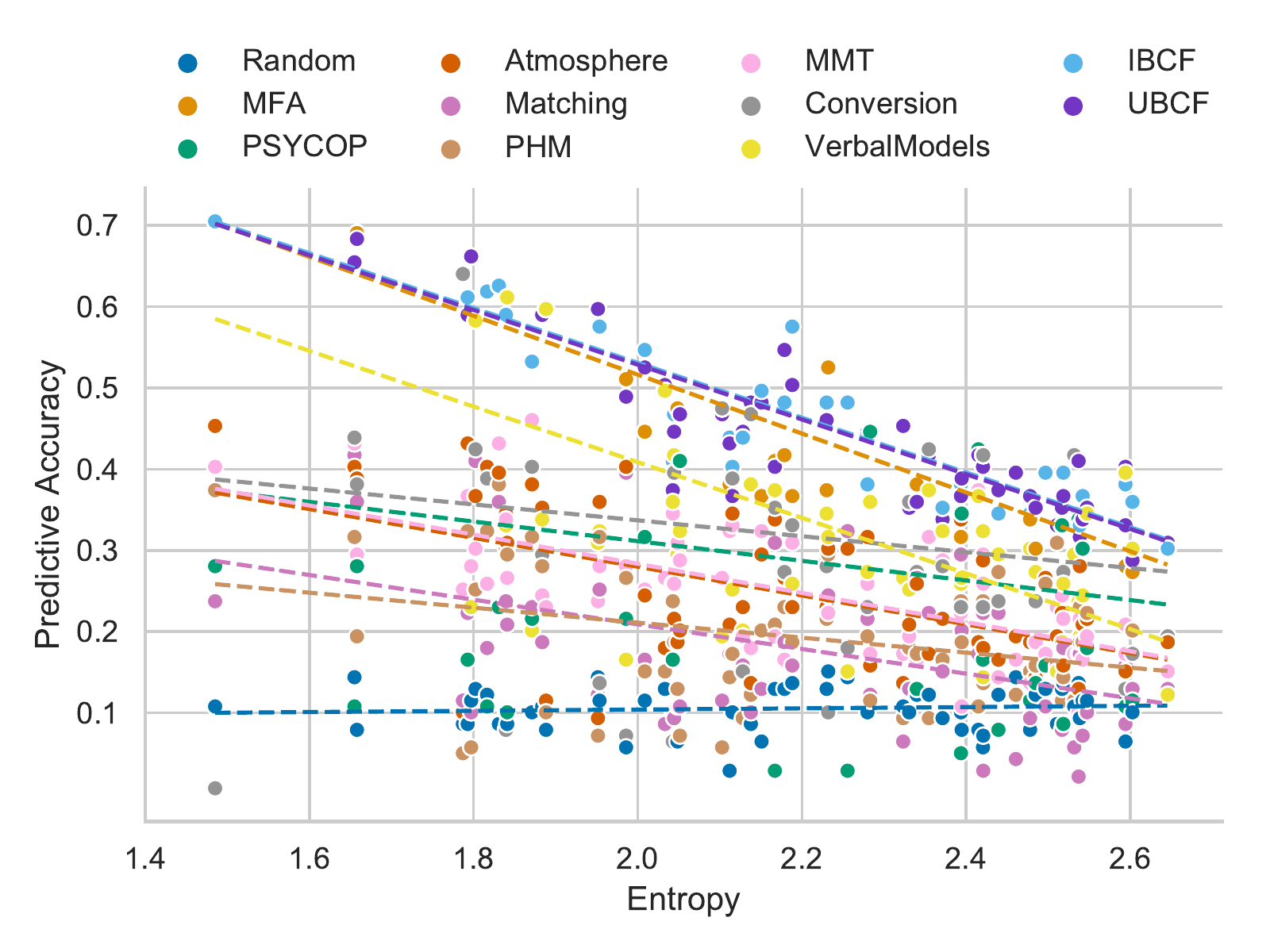}
        \caption{Relationship between syllogistic problems of varying entropy and model performances. Dotted lines represent interpolations between the data points.}
        \label{img:entropy}
    \end{figure}

    Figure~\ref{img:entropy} depicts the entropies of syllogistic problems with corresponding model performances. It shows a distinct gap in performance between the recommenders (IBCF, UBCF) and the remaining models. For low entropies, the recommenders are able to leverage the information encoded in the data resulting in high predictive accuracies. For higher entropies they are unable to maintain their initial distance to the cognitive models which are much more stable overall.

    Entropy in reasoning data can originate from (i) inconsistencies in the response behavior of individual human reasoners or (ii) interactions between independent reasoning strategies. The former point is a general issue of psychological and cognitive research since human participants are prone to lose attention due to boredom or fatigue. As a result, inconsistent and even conflicting data of single individuals can emerge \cite{Ragni2018}. Especially for collaborative filtering-based models this introduces substantial problems since users might not even be useful predictors for themselves. The latter point is a core challenge of cognitive science. Since reasoners differ with respect to their levels of education and experience with the task \cite{Nehrke1972}, recorded datasets are likely to be the result of a large number of individual strategies. For modeling purposes, the implications of both points differ greatly. Since inconsistencies due to lack of attention lead to behavior similar to guessing, it is unlikely for models to capture these effects by relying on behavioral data alone. Interactions between different strategies, on the other hand, are much more likely to be disentangled given additional insight into the domains and inter-individual differences between reasoners. Unfortunately, though, with the limited features currently contained in reasoning datasets, i.e., the responses, it is impossible to safely attribute the entropy of the data to either point. In the following sections, we therefore focus on collaborative filtering to shed light on the general capabilities of data-driven models in trying to uncover additional information about the problems of the domain.

\subsection{Strategy Simulation}

    Even though data-driven recommenders are able to achieve higher accuracies when compared to cognitive models, they are still far from perfectly predicting an individual reasoner. To investigate the remaining potential in the syllogistic domain, we need to gain an understanding of potential issues with the data.

    This second analysis considers artificial data with controlled levels of noise. Four of the cognitive models from the literature (Atmosphere, Matching, First-Order Logic, Conversion) were implemented and assigned to one of the four figures, respectively. By permuting the model-figure assignment and generating the corresponding response data we obtain $256$ artificial reasoners featuring interleaving strategies. The informativeness of this data is reduced by additionally introducing varying levels of random noise obtained from replacing conclusions with a random choice out of the nine conclusion options. With increasing levels of noise, the data should be less accessible for data-driven models.

    \begin{figure*}
        \includegraphics[width=0.49\textwidth]{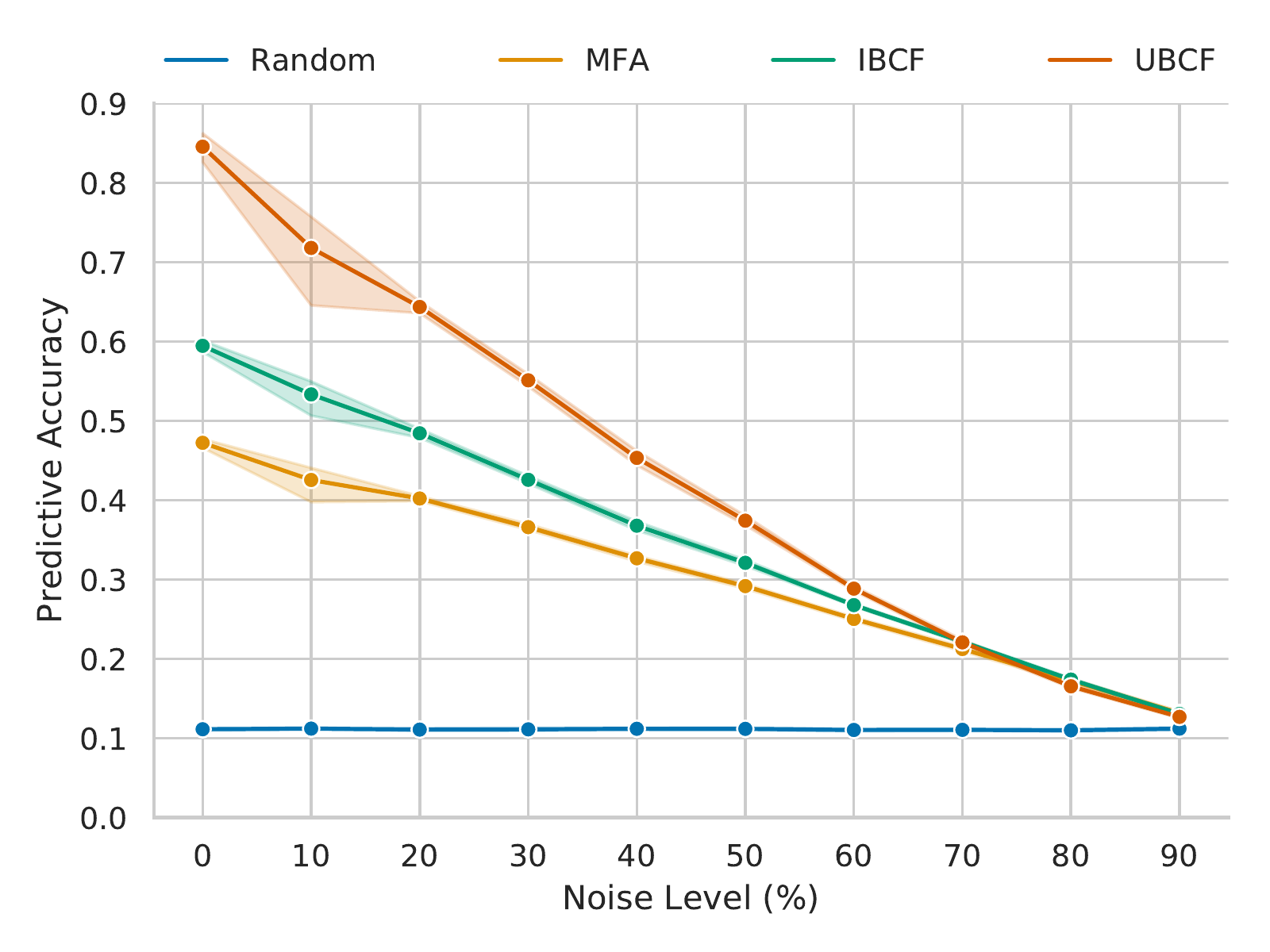}
        \includegraphics[width=0.49\textwidth]{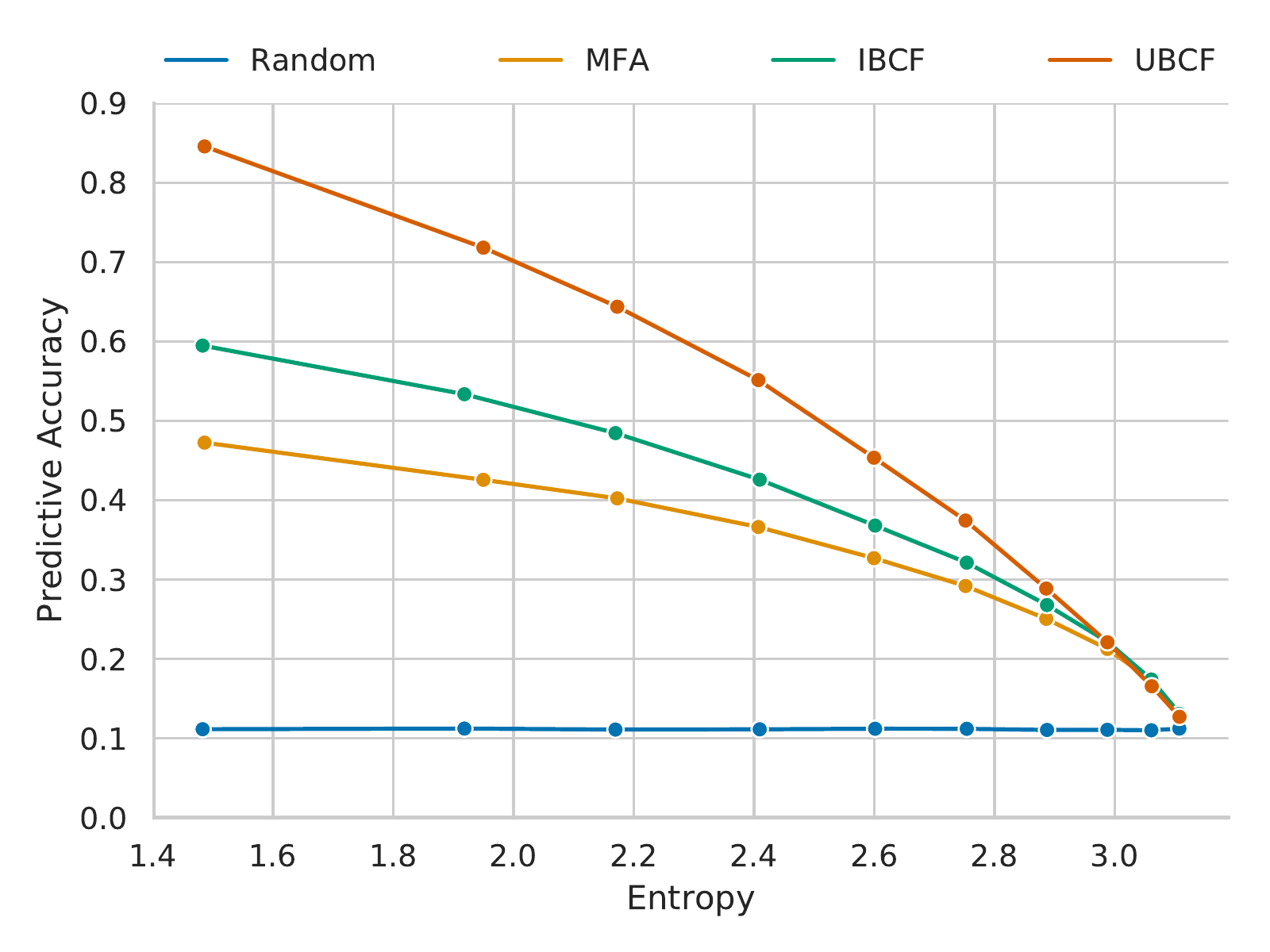}
        \caption{Strategy reconstruction performance of models based on artificial reasoning data with different levels of noise added by replacing a certain proportion of responses with a random choice from the nine conclusion options. The left and right images contrast performance with the raw noise proportions and entropies, respectively.}
        \label{img:boundsim}
    \end{figure*}

    Figure~\ref{img:boundsim} depicts the performance of the baseline and data-driven models on the artificial data. The left image plots the different noise levels against predictive accuracies. It shows that a decrease in response consistency has drastic effects on the models' capabilities to correctly predict responses due to the lack of information contained in the training data. The nearly linear relationship between the levels of noise and performance suggests that the models are stable in performance given the amount of reconstructable information. Consequently, they allow for a data-analytic assessment of ``noise'' in the data they are supplied with. In the case of syllogistic reasoning this means that close to $50\%$ of the data would effectively be indistinguishable from random noise. Explanations for this could be numerous ranging from too little data with respect to the number of possible reasoning strategies, over a lack of descriptive features, to guessing-like behavior, i.e. strategy-less decision-making on the side of study participants. The right image of Figure~\ref{img:boundsim} presents a different perspective on the impact of noisy data by computing corresponding entropies. Again, it shows that entropy is tightly linked to predictive accuracies.

    By comparison with the Figure~\ref{img:entropy}, some interesting conclusions can be drawn. In general, IBCF scores lower on the artificial data than on human data. Since IBCF is based on item-item dependencies, it is unable to directly exploit structural patterns of the data. It bases its predictions solely on information about ``reasoners responding x to problem A also respond with y to problem B''. Higher performance on the human data therefore suggests the existence of preferential clusters of reasoners which exhibit similar response behavior. Since the artificial data does not feature such groups but puts more focus on the structural information by evenly distributing the permutations of model-figure combinations, IBCF is at a disadvantage. While we cannot formally attribute the entropy observed in the human data to inconsistencies due to random noise, or varying overlap between distinct reasoning strategies, the properties of IBCF suggest the existence of key responses or groups exhibiting similar research patterns in the data which allow the method to perform some form of clustering to boost its accuracy. This can be interpreted as soft evidence for the second hypothesis, that the current problem with modeling syllogistic reasoning stems from the fact that features allowing for a disentanglement of strategies are scarce.

    A possibility to overcome these problems for the short-term progress of the field is by explicitly integrating assumptions about the structural properties of the data into models. If accurate enough, they should be able to boost models' capabilities to disentangle the overlapping strategies and allow for a general improvement of performance. Additionaly, the converse is true: if high-level theoretical assumptions lead to a significant improvement of the predictor, the theory is on the right track.

\subsection{Potential for Better Predictions}


    It appears as if a lack of information preventing the identification of strategies limits the potential of modeling in the domain of syllogistic reasoning. In general, there are two options to tackle this problem: improving models and extending the problem domain.

    There exist many possibilities to increase the predictive capabilities of models. On the one hand, additional features known for influencing reasoning patterns such as education \cite{Nehrke1972} or working memory \cite{Sues2002} can be integrated into the data to boost a model's ability to identify patterns. On the other hand, the model can be supplied with background information about the problem domain. Since cognitive science has a history of in-depth data analysis there is a lot of potential for integrating theoretical findings into models. We propose the use of collaborative filtering as an accessible tool for cognitive scientists to transform abstract insight into testable models.

\begin{figure}
    \centering
    \includegraphics[width=\columnwidth]{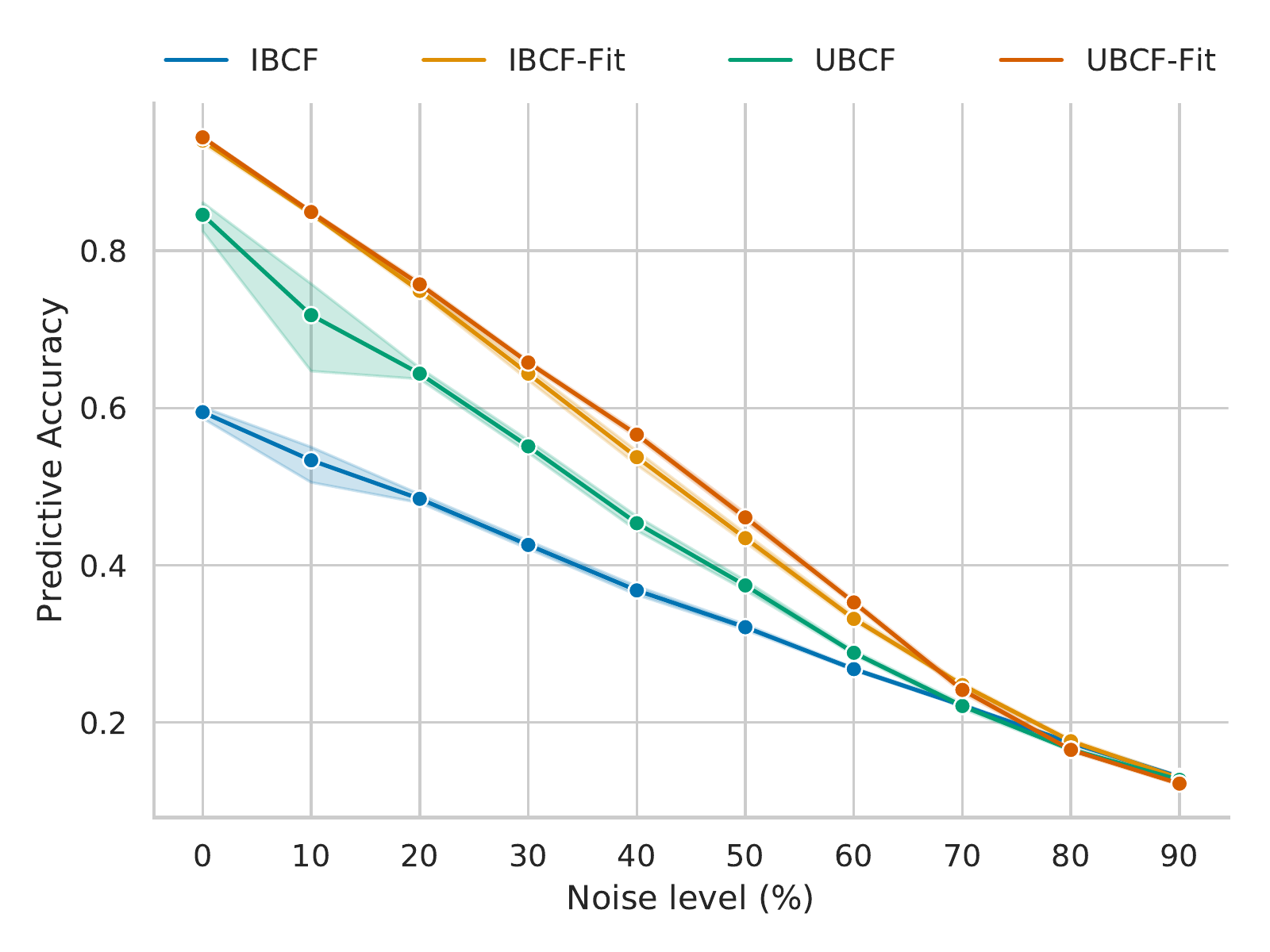}
    \caption{Comparison of item-based collaborative filtering (IBCF) and user-based collaborative filtering (UBCF) variants on artificially generated reasoning data. Fit-versions denote implementations where structural properties of the artificial data were actively integrated.}
    \label{img:cffit}
\end{figure}

    Figure~\ref{img:cffit} illustrates the potential of recommenders for insight-driven research by contrasting item-based collaborative filtering (IBCF) and user-based collaborative filtering (UBCF) with variants of them tuned to the structure of the artificially generated data, i.e., the observation that syllogisms of the same figure rely on the same inference mechanism. The plot highlights that this additional information about the data is able to push both IBCF and UBCF far beyond their initial performance. Especially for IBCF, the explicit integration of the structural foundation of the data lifts its performance to the same levels of UBCF. The gap between the domain-agnostic and tuned variants remains clearly visible even for high levels of noise. Even though explicit information about the structure of human data can only be approximated from theoretical insight into the domain, this shows that recommenders would be a useful tool for assessing the quality of assumptions.

    The second option to improve modeling of human reasoning is to extend the domain in question. If information about individuals is accumulated even across the borders of reasoning domains, models have more data to recognize descriptive patterns in. Additionally, it is possible to include distinctive background information about individuals such as personality traits. This approach has proven to boost performance in recommendation scenarios before and is likely to generalize to the reasoning domain \cite{Hu2011,Nalmpantis2017}. However, since the extension of the domain is out of scope for this work, we leave this idea open for future research.

    For research on human reasoning this final analysis shows that there exist data-driven methods which benefit from the integration of the kind of information that is usually uncovered in cognitive science and psychology. By integrating correlative insight into these kinds of models, the value of the findings can be directly assessed in benchmarking evaluations. Paired with more informative problem domains obtained from a unification of multiple domains of reasoning, or the addition of personality features about individual reasoners, data-driven and theory-driven research can collaborate to overcome the distance between the current state of the art and the goal of human-level AI.

  \section{Conclusion}
      Cognitive models for human syllogistic reasoning achieve unsatisfying accuracies when applied in a prediction setting. While the reasons for this could be numerous, it is interesting to see that data-driven recommenders based on collaborative filtering do not perform substantially better on an absolute scale. This raises concerns about the data foundation of reasoning research which is usually composed solely of reasoning problems along with the corresponding human responses.

Our results obtained from comparison with artificially generated data suggest that data-driven models are unable to identify and successfully exploit patterns in the structure of human reasoning datasets when, in theory, they should be able to. The two most likely explanations for this are noise in form of inconsistencies in the response behavior of humans, or a lack of distinctive features preventing data-driven approaches to identify the patterns required for successful predictions.

In order to advance the predictive performance to levels which are relevant for applications in the era of human-level AI, reasoning research needs to address its current shortcomings. Potential solutions include the improvement of models by a better integration of domain-specific insight as well as an active consideration of inter-individual differences, and the extension of the task for example by including other domains of reasoning, recording more comprehensive datasets, and leaving behind the current focus on data aggregation.

For integrating insight into models, we propose collaborative filtering recommenders as a general-purpose research method. On a technical level, they are easy to implement and understand, and outperform the current state of the art even in their domain-agnostic form. By integrating additional information about the domain (even if just on the level of correlations by weighting the dependencies between different features of the data), they allow for a transformation of abstract hypotheses into testable assumptions for modeling. Consequently, recommenders exhibit useful properties with respect to comprehensibility, especially in contrast to other methods from machine learning such as neural networks. As an example, they can naturally be applied to clustering contexts where stereotypical users are sought after.

Generally, we see a need for an increased focus on predictive accuracies for individual reasoners to allow more comprehensive benchmarking, to allow for a more accessible interpretation of the results, and ultimately to enable the models for real-world application. To facilitate this shift in perspective for other researchers, we released the tools driving our predictive analysis as a general benchmarking framework\footnote{\texttt{https://github.com/CognitiveComputationLab/ccobra}}. Only if the different disciplines of cognitive science find together to compete in modeling on unified informative domains using expressive and standardized metrics such as predictive performance, will human reasoning enter a level of progress relevant for human-level AI applications.

  \ack This paper was supported by DFG grants RA 1934/3-1, RA 1934/2-1 and RA 1934/4-1 to MR.

  \bibliography{recommenderanalysis}
\end{document}